\pdfoutput=1

\documentclass[11pt]{article}

\usepackage[preprint]{acl}

\usepackage{times}
\usepackage{latexsym}

\usepackage[T1]{fontenc}

\usepackage[utf8]{inputenc}

\usepackage{microtype}

\usepackage{inconsolata}

\usepackage{graphicx}
\usepackage{multirow}
\usepackage{booktabs}
\usepackage{csquotes}
\usepackage{enumitem}

\usepackage{todonotes}

\title{FRIDA to the Rescue! Analyzing Synthetic Data Effectiveness in Object-Based Common Sense Reasoning for Disaster Response}

\author{%
\textbf{Mollie Shichman$^1$,
Claire Bonial$^2$,
Austin Blodgett$^2$,
Taylor Pellegrin$^3$,} \\
\textbf{Francis Ferraro$^4$, Rachel Rudinger$^1$}
\\
$^1$University of Maryland, College Park, 
$^2$Army Research Lab \\
$^3$Oak Ridge Associated Universities,
$^4$University of Maryland, Baltimore County \\
\texttt{mshich@umd.edu},
\texttt{claire.n.bonial.civ@army.mil},\\
\texttt{ferraro@umbc.edu}, \texttt{rudinger@umd.edu}
}

\begin{document}
\maketitle
\begin{abstract}
During Human Robot Interactions in disaster relief scenarios, Large Language Models (LLMs) have the potential for substantial physical reasoning to assist in mission objectives. However, these reasoning capabilities are often found only in larger models, which are not currently reasonable to deploy on robotic systems due to size constraints. To meet our problem space requirements, we introduce a dataset and pipeline to create Field Reasoning and Instruction Decoding Agent (FRIDA) models. In our pipeline, domain experts and linguists combine their knowledge to make high-quality, few-shot prompts used to generate synthetic data for fine-tuning.  We hand-curate datasets for this few-shot prompting and for evaluation to improve LLM reasoning on both general and disaster-specific objects. We concurrently run an ablation study to understand which kinds of synthetic data most affect performance. We fine-tune several small instruction-tuned models and find that ablated FRIDA models only trained on objects' physical state and function data outperformed both the FRIDA models trained on all synthetic data and the base models in our evaluation. We demonstrate that the FRIDA pipeline is capable of instilling physical common sense with minimal data.
\end{abstract}

\section{Introduction}




\begin{figure}
    \centering
    \includegraphics[width=0.85\linewidth]{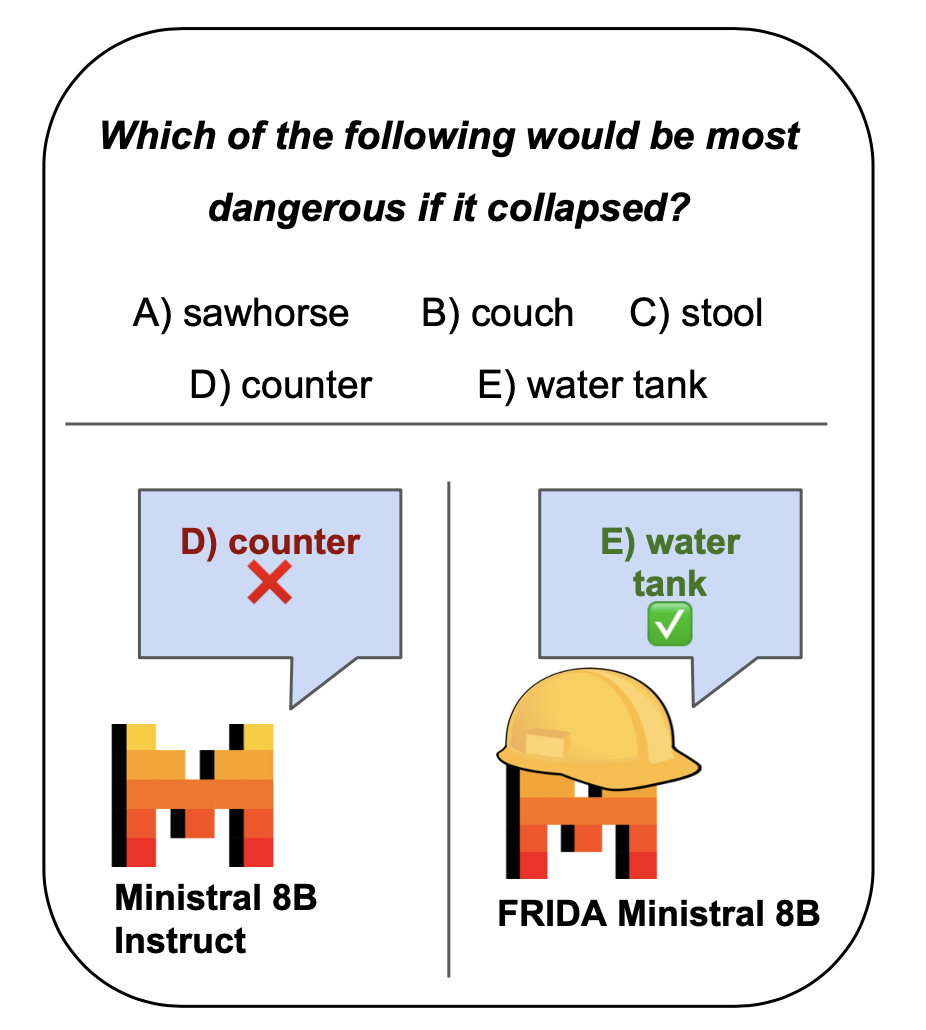}
    \caption{An example of how a FRIDA-tuned LLM outperforms its base model on questions combining an object's affordances and physical characteristics.}
    \label{fig:frida_ex}
\end{figure}
\textit{Which of the following would be most dangerous if it collapsed?} This question, as seen in Figure 
\ref{fig:frida_ex}, is fairly trivial for humans to answer, but requires several types of semantic knowledge. 
One must 
know the general size of these items and their other functions to fully assess the danger the item poses. A collapse is also a change of state that fundamentally shifts the use of these objects; a collapsed chair could be more likely to cut or scrape someone, but it could also mean the chair can now be carried if the chair folds. All of this knowledge is needed to answer this question, and all of it is embedded in our semantic understanding of objects that can cause danger and objects that can collapse, both intentionally and unintentionally.  

The ability to reason about objects is especially important in the context of human-robot interaction in disaster relief scenarios \cite{scout-justification}. For example, during search and rescue after an earthquake, a robot needs to know how to navigate partially collapsed buildings and how to use the many tools required to free people from the rubble. However, using robots to aid in disaster relief introduces many constraints. Because of the destruction a disaster can wreak, consistent internet connectivity cannot be assumed. For human safety, robots must be handled via radio in a secure location. This low-bandwidth communication means limited image data can be transmitted to the handlers, which rules out remote piloting \cite{scout-justification}. \textbf{We therefore need an autonomous system that can reason about its environment and the relief tasks required.}

As LLMs have improved dramatically, their abilities at semantic reasoning about objects have improved as well. LLMs have long been proven able to encode physical world knowledge \cite{knowledgebase}, and their embeddings can improve physical understanding of an environment and its objects both within and beyond a fine-tuned domain \cite{cohen2024surveyroboticlanguagegrounding}. 

However, much of this improvement is found only  
in larger models trained on more data \cite{wei2022emergentabilitieslargelanguage, scalinglaws}. This makes these essential semantic capabilities less accessible to our use case. Our robot cannot rely on an internet connection to make API calls. We instead must utilize the robot's limited on-board computing power, which can be as little as 16 GB of virtual RAM on an array of GPUs \cite{phil-warthog-specs}. That amount of GPU RAM can only reasonably run inference on a 13 Billion parameter model given the heuristics described in \citet{transformer-math-eleutherai}. Furthermore, this heuristic assumes that our robot is not running other processes in parallel, which is fairly unreasonable. We thus wanted to answer: \textbf{Given our constraints, how can we imbue all the physical common sense and semantics needed for smaller LLMs to be more capable at understanding a disaster environment?}


To answer our research question, we first tested the effectiveness of fine-tuning smaller models on disaster relief data. However, available data proved to be an additional constraint. Most publicly available data on disasters is social media-based reactions \cite{LLMdisaster}, which do not pertain much to our subdomain of disaster relief efforts. Furthermore, the specific knowledge (and to a lesser extent, the general knowledge) required for each mission varies by disaster. For example, after an earthquake, a robot needs to find survivors, while after a chemical spill, a robot needs to sample the environment for hazardous materials. \textbf{Therefore, we need a method for generating training data for specific disasters, and we need to evaluate which data are most effective at improving robot performance.} 

\begin{figure*}
    \centering
    \includegraphics[width=\linewidth]{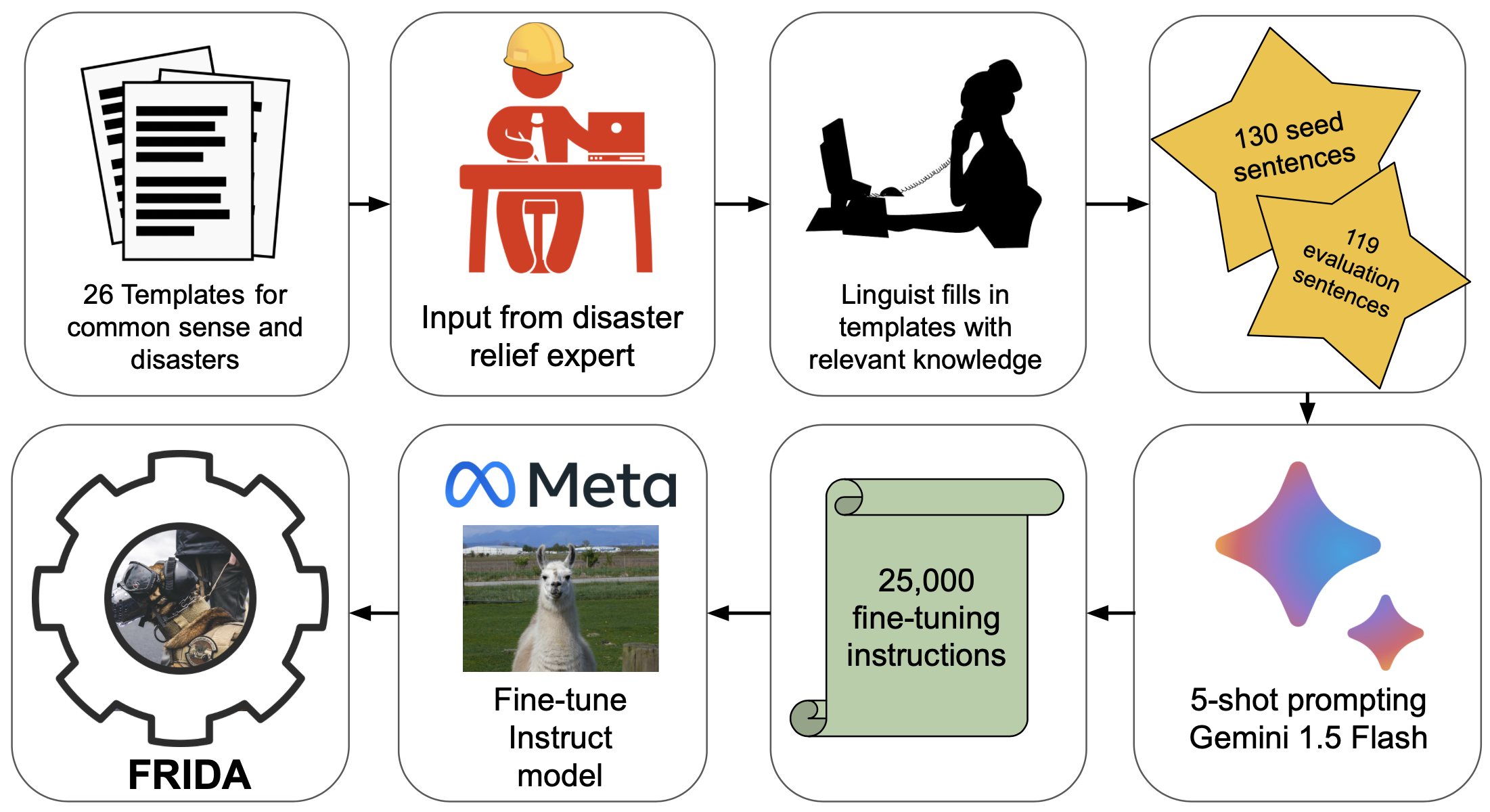}
    \caption{The pipeline to create the FRIDA suite of models. A search and rescue expert fills out a survey on the relevant tasks and objects used in disaster response, 
    then a semantics expert adds those terms to the ontology and fills in the templates to generate new seed instructions for a variety of different disasters. These seed sentences are utilized to generate synthetic data for fine-tuning an LLM with the necessary expertise on the specific disaster.}
    \label{fig:pipeline}
\end{figure*}

We present a pipeline to create Field Reasoning and Instruction Decoding Agent (FRIDA) models as a proof of concept for LLM viability in the disaster relief domain. For FRIDA, we leveraged both disaster and linguistic expertise to create gold-standard instructions that, in turn, are used as a basis for synthetic data generation, as seen in Figure \ref{fig:pipeline}. These synthetic data are then used to fine-tune smaller models that fit our memory constraints. Like its rescue dog eponym,\footnote{\url{https://en.wikipedia.org/wiki/Frida_(dog)}} our FRIDA models were initially developed and tested for earthquake disaster relief, based on expert knowledge pertaining to the February 6th, 2023 earthquakes in Turkey and Syria \cite{reuters}. \textbf{Thus, the resulting models are small enough to effectively operate onboard a robot and are fine-tuned on specialized and inexpensive data, satisfying all of our use case constraints.}

To investigate which synthetic data most influenced model performance, we ran an ablation study where we fine-tuned the same small LLMs on subsets of our synthetic data corresponding to specific types of object-based reasoning. We call these resulting models the ablated FRIDA (aFRIDA) models. We found that aFRIDA models trained on  general semantics and physical common sense had stronger overall performances than models trained on only domain-specific knowledge. Additionally, the best performing aFRIDA models scored better than their corresponding base models and FRIDA models trained on the entire synthetic dataset. We posit that FRIDA succeeds in improving object-related general common sense, but that small LLMs struggle with disaster-specific equipment usage.

Our contributions are as follows:
\begin{enumerate}
    \item An expert-in-the-loop pipeline (Figure~\ref{fig:pipeline}) 
    for generating specific and high-quality synthetic data that can be used for fine-tuning when man-made data are not feasible to obtain, as well as the resulting gold-standard datasets.
    \item A synthetic dataset of 25,000 instructions relating to object reasoning and earthquake response with accompanying analysis.
    \item The FRIDA model, fine-tuned on Mistral AI's Ministral 8B model with the above synthetic data, which investigates small LLM potential. 
    \item A series of ablated FRIDA (aFRIDA) models trained on subsets of the synthetic dataset to investigate which synthetic data were most effective.
    \item An in-depth analysis investigating the challenges of imbuing physical common sense and complex object reasoning into LLMs. 
\end{enumerate}
Our datasets, code, and a complete walkthrough of the FRIDA pipeline are currently available.\footnote{\url{https://github.com/mshich1/FRIDA/}}

\section{Related Work}
\subsection{LLMs Reasoning about the World}
There are a wide variety of methods for leveraging LLMs for reasoning in a physical environment based on Chain of Thought prompting \cite{CoT}. These include variants like re-prompting \cite{reprompt}, which prompts the LLM to regenerate a plan if certain criteria aren't met at certain steps, or Tree of Thought \cite{ToT}, which generates a tree of potential steps and evaluates each potential path via either a breadth-first or depth-first search. 

There are also methods that allow the LLM to take in environmental feedback in response to its output. For Inner-Monologue \cite{innermono}, the LLM is given the option to ask for more scene descriptors from a human handler, which it then incorporates into its prompts, improving task completion and decreasing hallucination. Another example is SayPlan \cite{sayplan}, which uses 3D scene plans to iterate on proposed strategies until an effective path is discovered. \citet{humanplan} get feedback from LLMs themselves by using a wide variety of LLM agents to do various sub-tasks for planning, including generating a general outline, using external tools to gain information, and evaluating which plan is best. 

One resource for improving LLM understanding of an object's functions, also known as the object's affordances, is \citet{Text2Afford}, who curate a dataset of naturally occurring sentences and corresponding images. They then transform them into inference, probing, and masking tasks for LLMs and Visual Language Models (VLMs). Their evaluation shows that VLMs do not have straightforward understandings of object affordances, but few-shot fine-tuning improves LLM and VLM performance on identifying object affordances. This work focuses on building a stronger basis in LLMs to improve these downstream tasks, as well as understand which data are most important for a robot's success.

\subsection{Disaster Work and Natural Language Processing}
\citet{LLMdisaster} completed a systematic search and analysis of over 100 peer-reviewed papers relating to Natural Language Processing (NLP) tools being applied to disasters. 85 of the 107 papers found were analyzing social media, and the majority of papers focused on sentiment analysis, text classification, and information extraction tasks. Both the data sources and NLP tasks do not have a clear parallel with our objective.

While robots have been successfully deployed in disaster relief missions, the current state of the art is a human tele-handler in complete control of the robot \cite{controller-sota,vr-controller-sota}. This puts all of the cognitive burden on said tele-handler, and does not allow for the re-tasking and pivoting required in such a high-stakes, fast changing scenario \cite{scout-justification}. To move the state of the art from tele-handling to human-robot dialogue, \citet{scout-corpus} provide a corpus of simulated dialogues in a disaster scenario that are annotated for semantic meaning, dialogue structure, and visual common ground. However, this corpus works with a robot with limited abilities and does not touch on creating a system to reason about a wide variety of objects and disasters.

\subsection{Synthetic Data Generation}
Synthetic data, or data generated by an LLM, has become increasingly popular as an inexpensive and relatively proficient method of data collection. While cyclically fine-tuning LLMs on the synthetic data they generate denigrates the models' performance \cite{MAD}, fine-tuning on synthetic data has nevertheless improved short term performance in instruction following and social common sense \cite{tinystories,self-instruct}. 

This paper is inspired in particular by the pipeline developed by \citet{self-instruct}, who hand crafted 175 ``seed'' instructions. These seed instructions were used for 8-shot  prompting of GPT's \texttt{text-davinci-001} model to generate more than 50,000 instructions for a generic and ungrounded AI assistant. These synthetic instructions were then used to fine-tune \texttt{text-davinci-001}. The authors found that their method and resulting fine-tuned model performed comparably to OpenAI's GPTInstruct \cite{self-instruct}. \citet{alpaca} innovated on \citet{self-instruct} by fine-tuning a separate, smaller language model with a different architecture, as opposed to fine-tuning on the same model that generated the data. They subsequently found that their resulting model's answers were rated as highly as GPT's \texttt{text-davinci-003}. 


\section{Methods}
\subsection{
FRIDA Seed Data}
\label{sec:data}
We developed an expert-in-the-loop pipeline to generate high-quality seed data that leverage expertise on both disaster-relief and semantics. The purpose of this pipeline is to enable quick and efficient fine-tuning of small LLMs to be capable of critical reasoning in specific disaster environments.  
The details of this pipeline are described in \citet{ppdc}, here we provide a brief overview.  
We developed a series of templates that can be filled in with vocabulary from an affordance ontology based on the Rich Event Ontology \cite{kazeminejad2018automatically}. This affordance ontology is extended to serve as 
 an ontology of disaster-related objects and their functionalities, as defined by the objects' PropBank semantic roles labels \cite{palmer2005proposition}. 
 
To fill in these templates with proper data, a disaster expert first provides information about the relevant objects and situations encountered in their work. For this paper, the authors simulated this step by gathering existing resources authored by experts on the Turkey-Syria Earthquake recovery efforts \cite{reuters}. After gathering domain-specific data, linguists go through a template-filling pipeline. Summarily, the linguists select the relevant vocabulary from the expert knowledge to add to the aforementioned affordance ontology. They then use this ontology and template-specific generation instructions to fill in the templates to create 
``seed'' instructions. These templates are formatted as multiple choice questions with semantically distinct answers. Some examples of this process, as well as some of the synthetic instructions that result, can be seen in Table \ref{tab:data_ex}. 

\begin{table}[t]
    \centering
    \begin{tabular}{p{1.75cm} p{5cm}}
    \toprule
     Template &   What state should \textbf{OBJECT} be in to easily use it: \textbf{X STATE} or \textbf{Y STATE}? \\
     \cmidrule(lr){1-1}\cmidrule(lr){2-2}
     Seed Instruction & What state should a \textbf{drawbridge} be in for cars to cross a river? \textbf{A) Lowered} or \textbf{B) Raised}  \\ 
     \cmidrule(lr){1-1}\cmidrule(lr){2-2}
     Synthetic Instruction& What state should a \textbf{door} be in to easily enter a room? \textbf{A) Open} \textbf{B) Closed} \\
     \toprule
     Template & What role does \textbf{OBJECT} play in \textbf{DISASTER-RELATED TASK}\\
     \cmidrule(lr){1-1}\cmidrule(lr){2-2}
     Seed Instruction & What role do \textbf{hydraulic lifts} play in \textbf{rescuing people after an earthquake?} \\
     \cmidrule(lr){1-1}\cmidrule(lr){2-2}
     Synthetic Instruction& How is a \textbf{crowbar} typically used during \textbf{earthquake rescue operations}?\\
     \bottomrule
    \end{tabular}
    \caption{Two Examples of templates and their corresponding gold standard and synthetic instructions. Note that the blanks in the first template can only be filled in by objects with multiple states (i.e. linguistic knowledge), while the blanks in the second template can only be filled in with specific tools (i.e. disaster expert knowledge).}
    \label{tab:data_ex}
\end{table}

Although some related work leverages the same seed sentences used for generating synthetic data to also evaluate the data \cite{self-instruct}, we used this same pipeline to develop a separate and unique evaluation to ensure that our evaluation was not present in any training data. The seed and evaluation instructions include multiple choice answers, enabling more efficient evaluation and comparison of models. 

\subsection{Synthetic Dataset Generation and Analysis}
\label{sec:generation}
\begin{figure}
    \centering
    \includegraphics[width=\linewidth]{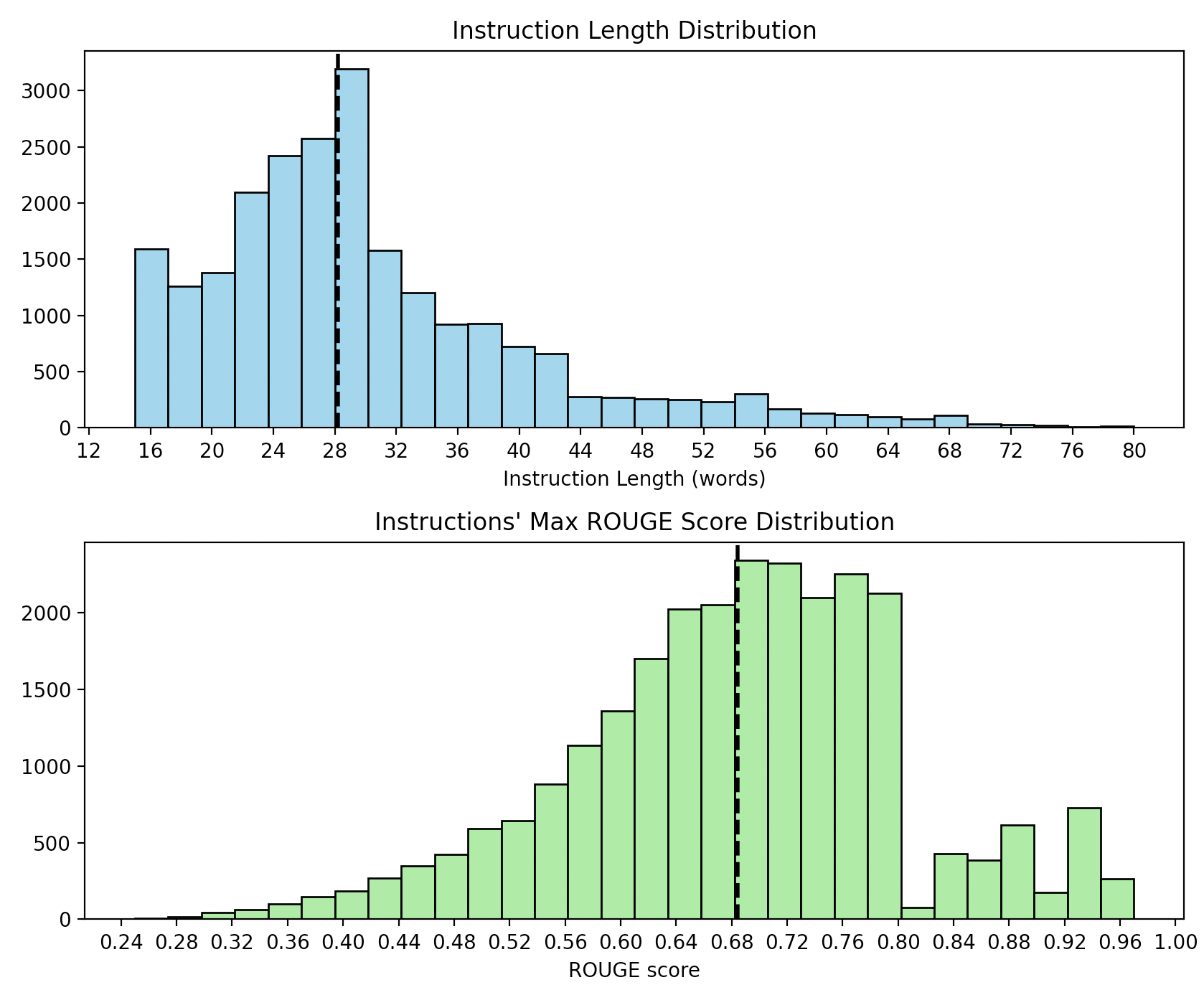}
    \caption{The distribution of the synthetic data's instruction length (top) and maximum ROUGE score (bottom). Averages are shown as black dashed lines. Our high average instruction length and general distribution shows synthetic instructions are sufficiently complex, and our average over each instruction's top ROUGE score shows the instructions are sufficiently unique for this to be a challenging task.}
    \label{fig:hist}
\end{figure}
The dataset we use in this work focused on search and rescue operations in the aftermath of the Turkey-Syria Earthquake \cite{reuters}. We had 26 templates grouped into 8 categories based on the type of knowledge they query as defined by the Generative Lexicon Qualia \cite{gen-lexicon}. For all categories and examples, see Table \ref{tab:templates} of Appendix~\ref{appendix:categories}.
For each template, expert annotators hand-made 5 seed instructions for synthetic data generation (130 total instructions) and a minimum of 4 evaluation instructions (119 examples). All resulting instructions were examined by a second author for correctness. 

For each template, we used its corresponding seed instructions for 5-shot prompting with Gemini-1.5-flash to generate 980 synthetic instructions based on the given template \cite{gemini}. We chose Gemini as our synthetic data generator for its accessible and affordable API, as well as its high scores on our evaluation (93.9\% average Semscore, see section \ref{sec:eval}). We prompted Gemini to return 40 instructions per API call. To ensure our synthetic data were unique, we used ROUGE scoring \cite{rouge} to ensure Gemini was not giving us duplicates of previously generated instructions. Depending on the template, the cut-off ROUGE score went from 0.8 for templates with more varied language to 0.97 for templates with very structured wording. We also increased model temperature for the more structured templates to increase diversity of responses.

We get a sense of the resulting synthetic dataset from the histograms in Figure \ref{fig:hist}. We automatically evaluated for instruction length and each instruction's maximum pairwise ROUGE score. We found we had substantial average instruction length, and reasonable ROUGE scores given that our data are template-based. There was a large range in both metrics across the different template categories, which we attribute to the overall complexity of the individual templates. Some templates require short instructions with binary answers, while others have longer instructions where all answers are sentences. 

\begin{table}
    \resizebox{\columnwidth}{!}{
    \centering
    \begin{tabular}{l r}
    \toprule
    Category&Training / Dev split\\
    \midrule
     Relative Size & 3620 / 403 \\
     Object Functions & 4460 / 496 \\
     Objects Causing Harm & 2675 / 298 \\
     Earthquakes & 882 / 99\\
     Specialized Equipment & 2679 / 298 \\
     Instruction Understanding & 1792 / 200 \\
     Differences & 4458 / 496 \\
     Non-functional Object Facts & 2662 / 296 \\
     \midrule
     Total Instructions & 23232 / 2582\\
     \bottomrule
    \end{tabular}
    }
    \caption{The number of instructions in the training and development datasets used for fine-tuning FRIDA and its ablations.}
    \label{tab: set_size}
\end{table}
\subsection{FRIDA Model construction}
We used our synthetic dataset to fine-tune the 1 Billion, 3 Billion, and 8 Billion parameter Instruct models from the LLaMa 3 herd \cite{llama3} as well as the Mistral AI's Ministral 8B Instruction tuned model \cite{mistral7}. We chose to use the LLaMa suite due to it having multiple small instruction tuned models of different sizes, with strong performance \cite{llama3}. We chose Ministral 8B to serve as a comparison, since it is trained with sliding window attention, unlike the LLaMa models trained with full attention \cite{mistral7}. Additionally, Ministral 8B was released after the LLaMa 3 herd and outperformed the LLaMa models on a variety of metrics \cite{mistral7}. We chose to fine-tune the instruct variations of these models because our task is based on answering questions. All models were trained with the performance enhanced fine-tuning model LoRA \cite{lora} with full precision. 

Of the four fine-tuned models, the strongest fine-tuned model performance on our evaluation was from models trained on Ministral 8B. We hypothesize that this is due to the architectural differences between Mistral AI and Meta AI models. Specifically, sliding attention could be helpful in focusing the model's attention on the instruction content instead of the multiple choice answers. Additionally, Ministral 8B's sliding attention mechanism is more memory and time efficient, making it more practical for deployment on a robot \cite{mistral7}. As such, we focused our analysis on FRIDA and aFRIDA models based on Ministral 8B, since they are the most conceivable models to work in a robotic system in the near term. Results for the LLaMa models can be found in our github.
Fine-tuning specifics can be found in Appendix \ref{sec:ft}.

\subsection{Evaluation}
\label{sec:eval}
As described in section \ref{sec:generation}, we used the same pipeline for creating seed data to create a custom evaluation, with at least four evaluation questions per template for a total of 119 evaluation instructions.

Although we leverage multiple choice questions and answers for evaluation, we required a less rigid method than exact match so that formatting errors (e.g., writing ``A'' instead of ``A)'', or forgetting punctuation) would have less impact. 
Thus, we used SemScore \cite{semscore,semcode}, which is a scoring metric that uses cosine similarity to compare a model's embedding vectors of the gold standard and FRIDA responses. 


\subsection{Ablation Study}
To better understand the effectiveness of the types of physical reasoning represented in our synthetic data, we ran an ablation study where we fine-tuned our base model on subsets of the synthetic fine-tuning data, which can be seen in Table \ref{tab: set_size}. 
We made an ablated model for each category of data, where each model is fine-tuned only on the synthetic data generated by templates in said category. For example, the ``Relative Sizes and Shapes'' ablation model is trained on data generated from 4 templates testing size, weight, objects fitting in containers, and objects changing state. We refer to these ablated models as \textbf{ablated-FRIDA} (or \textbf{aFRIDA}) models.

The resulting name for a FRIDA model trained only on data from the Relative Sizes and Shapes category would thus be, ``aFRIDA: relative sizes and shapes'', where ``relative sizes and shapes'' refers to the subset of data used (see Appendix Table~\ref{tab:templates} for data categories). The ablated models were tuned with the same hyper-parameters and hardware as the full FRIDA model.

A model suite for a given base model contains FRIDA, trained on the full dataset, as well as 8 aFRIDA models trained on the categorical subsets of the data: relative sizes and shapes, object function, object differences, specialized equipment, objects causing harm, non-function object facts, earthquake knowledge, and instruction understanding. Examples of data for each category can be found in Table \ref{tab:templates} in the appendix.
\section{Results}

\begin{table}[]
    \centering
    \begin{tabular}{l c}
    \toprule
        Model & SemScore\\
         &  Accuracy (\%) \\
    \midrule
        Ministral 8B Instruct & 93.5\\
        FRIDA & 94.6 \\
    \midrule
        Ablated Model& SemScore \\
        Fine-Tuning Data Subset & Accuracy (\%)\\
        \midrule
        relative sizes and shapes&  \textbf{95.0}\\
        object functions & 94.7\\
        object differences&93.4 \\
        objects causing harm&93.3\\
        specialized equipment& 93.8\\
        non-functional obj facts&93.2\\
        earthquake knowledge&91.7\\
        instruction understanding&85.0\\
        \bottomrule
    \end{tabular}
    \caption{The SemScore Accuracy on \textbf{all evaluation data} for the base model Ministral 8B Instruct, the fine-tuned FRIDA model trained on all synthetic data, and the fine-tuned models trained on ablated subsets of the synthetic data (aFRIDA). The FRIDA model trained on all data improved performance over its corresponding base model. The best overall performance came from the aFRIDA model trained on a subset of the synthetic dataset involving comparing objects by their physical state.}
    \label{tab:accs_for_all}
\end{table}
As seen in Table \ref{tab:accs_for_all}, the Ministral 8B FRIDA model had a higher 
SemScore Aacuracy than its base model. However, the aFRIDA models for the ``Relative Size and Shape'' and ``Object Functions'' categories outperformed both the unablated FRIDA model and the base model. These models also outperformed Gemini-1.5-flash's SemScore of 93.9 in a zero shot setting. 

\begin{figure}
    \centering
    \includegraphics[scale=0.28]{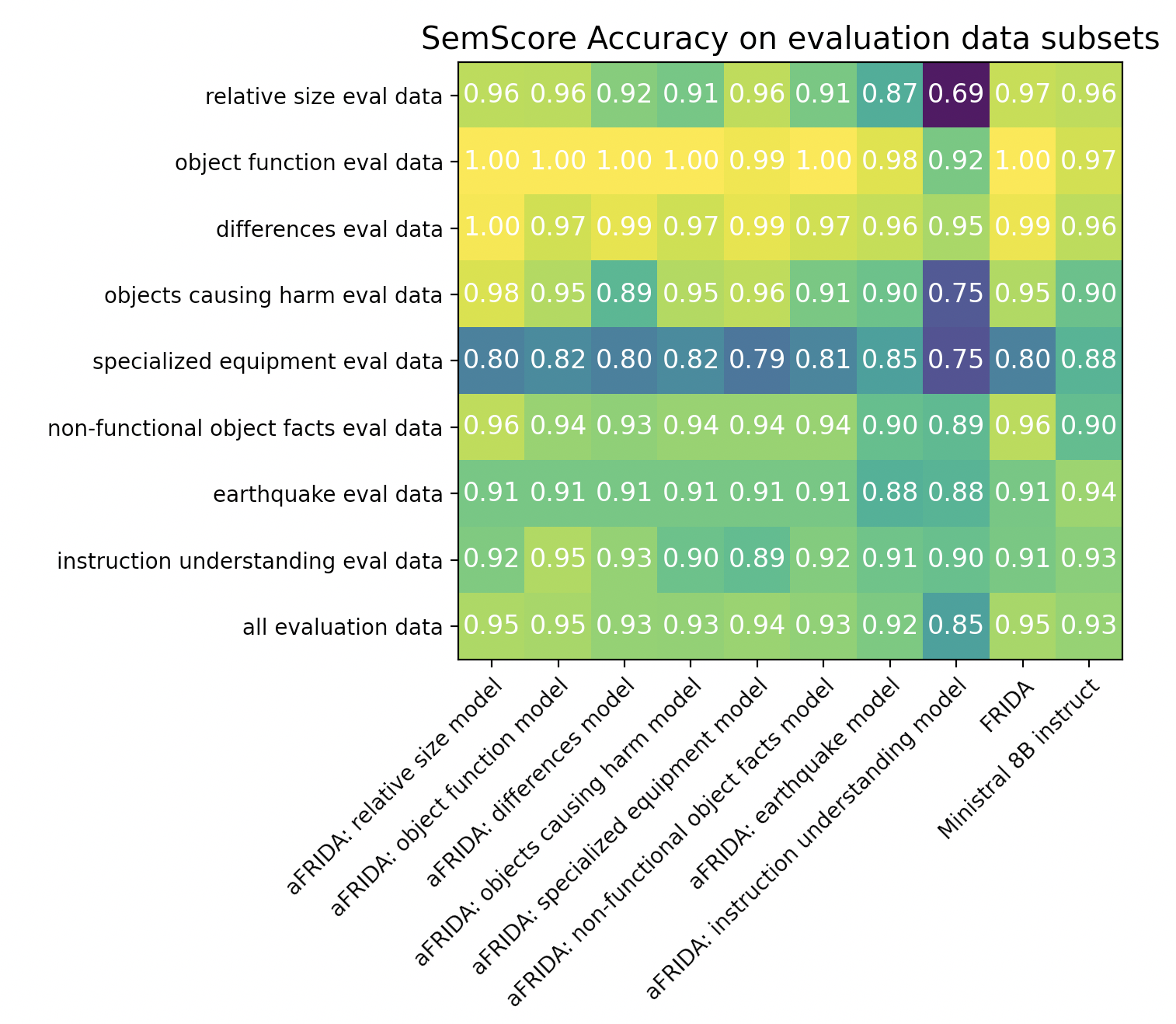}
    \caption{SemScores (embedding-vector cosine similarity scores) for the FRIDA suite for each type of evaluation. Across all models, performance is better in evaluation data corresponding to physical common sense (object functions, differences) and worse in evaluation data corresponding to specialized object knowledge (earthquake, specialized equipment). 
    }
    \label{fig:heatmap}
\end{figure}

We assessed each model's capability on each type of reasoning tested in the evaluation dataset. To show the overall trend across models, we present the SemScore results for the FRIDA and aFRIDA models in Figure \ref{fig:heatmap}. Overall, when observing model performance in the Figure \ref{fig:heatmap}'s columns, 
models fine-tuned only on objects' basic size and shape characteristics or only on object functionality performed more strongly across most evaluation categories. This was despite these synthetic data covering straightforward physical semantics that don't require any highly specific knowledge or creativity like the ``specialized equipment'' or ``objects causing harm'' categories. These models also had the strongest performance with far less training data than the full FRIDA model (see Table \ref{tab: set_size}).

Looking at evaluation data types represented in the rows, it is clear that the more difficult evaluations are ``specialized equipment'', the category querying about the specialized objects used in earthquake search and rescue, and ``earthquake'', the category evaluating scientific knowledge about earthquakes. Both of these evaluations are highly specific and technical. The easier evaluation categories are ``object functions'' and ``differences'', which pertain to understanding the basic semantics of objects' abilities and the differences between objects, respectively. 

Another key observation from Figure \ref{fig:heatmap} can be found by comparing evaluation performance between 
FRIDA and Ministral 8B. FRIDA has stronger performance than the base model except for the ``required equipment'', ``earthquake'', and ``instruction following'' evaluations. This could potentially demonstrate that these data need to be generated differently or that Ministral 8B needs more of them in order to strengthen performance. 

\section{Discussion \& Error Analysis}
It is particularly surprising that the ``aFRIDA relative size and shape'' and the ``aFRIDA object function'' models outperformed all other models across the board, even though the physical semantics expressed in those fine-tuning data are not complex. We hypothesize that clarifying the basic properties and affordances of objects provided a better basis for the model to have stronger physical reasoning across all categories. 

Another surprise was that the ``relative sizes and shapes'' evaluation subset was a challenge for the FRIDA suite. Although one may think that simpler object properties like its ``relative sizes and shapes'' might be relatively prevalent in the base models' pre-training data, it is also plausible that reporting bias in web text leads to under-representation of highly commonplace facts \cite{raji2ai}. We hypothesize that this lack of pretraining data is partially why the ablation model trained on ``relative shapes and sizes'' synthetic data performs so strongly. However, this does not answer why the ablated models trained on data pertaining to other challenging categories in our evaluation, namely ``aFRIDA: earthquake'' and ``aFRIDA: specialized equipment'', did not receive the same overall performance bump.

We suspect that the reason ``aFRIDA: earthquake'' and ``aFRIDA: specialized equipment'' did not similarly improve performance is that our synthetic data for the more specific objects and tasks tended to be longer and have lower ROUGE scores. These data therefore had more diversity. The sample size of the Earthquakes and Specialized Equipment synthetic data subsets may have been too small for the model to be correctly biased by fine tuning. Conversely, larger models may have ingested operators' manuals for specialized equipment, facilitating parroting answers for questions on this topic. We note that our research highlights the general difficulty of analyzing the precise effects of fine-tuning given opaque pre-training data. 

Error analysis of both the FRIDA model (fine-tuned using all synthetic data) and the ``aFRIDA relative size and shape'' model revealed that both models got the same instances and number of the ``relative size and shape'' evaluation data incorrect. For example: 
\begin{enumerate}
    \item ``What is the easiest way to use a camera?'' \\ 
    A) with the camera plugged in \\ 
    B) with the camera unplugged\\
    Gold: B) with the camera unplugged\\
    FRIDA: A) with the camera plugged in\\
\end{enumerate}
The base Ministral model generally gets the same ``relative size and shape'' evaluation instances incorrect as the FRIDA models. However, it also answers incorrectly for over half of the  instances of test items that relate to answering which item is bigger and which item will fit into another item. For example: 
\begin{enumerate}[resume]
    \item ``Choose the biggest of a given set of objects in terms of your own common sense.''\\ A) bicycle\\
    B) chalk\\ 
    C) poster\\ 
    D) jar\\
    E) taillight\\
   Gold: A) bicycle\\
   Ministral: D) jar\\
   \item ``Can chalk fit in a cup?'' \\
   Answer ``it can'' or ``it cannot''\\
   Gold: it can\\
   Ministral: it cannot\\
\end{enumerate}

Thus, we conclude that the fine-tuning contributed to improvement in understanding which items are bigger and which items fit into others in particular. This improvement may translate to improvement in other related categories. Specifically, we also see dramatic improvement over the base model for the ``objects causing harm'' evaluation data. This could be further boosted by a general understanding of which objects are larger. 

When it came to reasoning about the complex equipment used, error analysis revealed that both vanilla and fine-tuned models scored perfectly when asked to choose the correct role for an object in an event. For example:
\begin{enumerate}[resume]
    \item ``What role does a helicopter play in the search and rescue process?''\\
    A) Provide a vantage point to identify heavily damaged areas\\
    B) Move large vehicles to disaster area\\
    C) Blow away debris\\
    D) Warn victims about aftershocks\\
    E) Blow debris out of the way\\
    Gold: A) Provide a vantage point to identify heavily damaged areas\\
    Ministral: A) Provide a vantage point to identify heavily damaged areas\\
    FRIDA: A) Provide a vantage point to identify heavily damaged areas
\end{enumerate}

The task of choosing the correct object to use for a task proved more challenging. Fine-tuning on related data seemed to unnecessarily bias the model toward choosing the most complicated object, while fine-tuning on unrelated data maintained results. For example:
\begin{enumerate}[resume]
    \item ``Select the equipment needed for breaking rubble into smaller pieces after an earthquake.''\\
    A) axe\\
    B) pickaxe\\
    C) hydraulic lift\\
    D) hard hat\\
    E) hammer\\
    Gold: B) pickaxe\\
    Ministral: B) pickaxe\\
    FRIDA: C) hydraulic lift\\
    aFRIDA relative sizes: B) pickaxe
\end{enumerate}

In the most complex reasoning task of ordering steps to complete to use an object, fine-tuning had no clear effect, with all models providing random answers.

\begin{enumerate}[resume]
    \item ``The following are two different steps for using a dump truck. Which needs to happen first?\\
    A) Wait for others to fill the truck bed\\
    B) open the tailgate\\
    Gold: B) open the tailgate\\
    Ministral: B) open the tailgate\\
    FRIDA: B) open the tailgate\\
    aFRIDA relative sizes: A) Wait for others to fill the truck bed\\
    aFRIDA required equipment: A) Wait for others to fill the truck bed\\
\end{enumerate}

We thus conclude that fine-tuning for required equipment did not effectively bias the models to understand the use cases of these complex objects. At its worst, it incorrectly biases the model to choose complex objects when simpler ones would be more effective.

Overall, the FRIDA pipeline 
improves small LLM object reasoning when said models are fine-tuned on more general physical common sense and object reasoning data. The FRIDA suite models are lightweight enough to fit within our constraints, and can even achieve comparable performance to a much larger Gemini model. 
In comparison to the ablated models, the performance of the full FRIDA model trained on all synthetic data demonstrates that more work needs to be done to improve the synthetic dataset distribution to be ideal for improving FRIDA model performance on reasoning for earthquake search and rescue. 


\subsection{Future Work}
\label{sec:fw}
There are several ways we can further improve the FRIDA pipeline. We want to improve our prompting for synthetic data to make them less trivial to answer. We can refine and expand our less technical templates. By adding different phrasing, we hope to make our synthetic data more reflective of real world natural language. We also hope implementing the strategies in other work \cite{personas, ultrachat, orca} for diversifying synthetic data will improve generation quality and efficiency. We want to explore the impact of using quantized models over full precision models to determine if we can save additional storage space while maintaining reasoning ability. Finally, we plan to test the pipeline on other domains with experts to help us refine our process. 

\section{Conclusion}
We introduce a pipeline to create expert-in-the-loop-based synthetic data that is then used for fine-tuning to create FRIDA models. We found our pipeline improved performance over our base model. We performed an ablation study and found that data generated from templates based in basic physical common sense reasoning about objects improved performance most; ablated models trained on those data scored higher than FRIDA models trained on all synthetically generated data and higher than Gemini-1.5-flash, the LLM that generated the synthetic data. This pipeline is an important step in understanding and improving LLM object reasoning for practical use. Even if some of our problem constraints are eventually alleviated by technology that facilitates very large models with smaller compute requirements, there will remain problem spaces for which web-based pre-training data simply does not exist. Our research demonstrates an effective pipeline to specialize models fine-tuned on data that is not well-represented in typical web text pre-training data. 

\section{Limitations, Risks, and Ethics}
One limitation is that we train and evaluate on template-generated data rather than naturally occurring language; there could be linguistic or stylistic differences between template-generated data and naturally occurring instructions. Though our approach still relies on access to expert input and non-trivial computational power for fine-tuning to counter these shortcomings, we outline solutions in Section \ref{sec:fw} which we believe are ripe avenues for future work. 

We note that multiple choice questions can be different and less complicated than an unconstrained turn between a user and an AI assistant. Nevertheless, we believe this work is an important step towards our goal of imbuing smaller language models with physical common sense. This is because we prove the feasibility and capability of small LLMs to complete this more constrained task. We argue that FRIDA should be seen as a proof-of-concept for LLM physical common sense understanding, which sets the stage for increasingly challenging training data and evaluations.

FRIDA is built by biasing an LLM to a specific domain. While this is important for our work, this could be misused to bias models in harmful ways, especially when considering applications involving social common sense. When modifying our seed data and templates, we took care to reduce gender bias as much as possible. This was fairly trivial since all questions pertained to objects and events, not people. We acknowledge that many objects from the ontology we used were annotated with a Western perspective, and that other cultures likely have additional uses for these objects.
\bibliography{acl_latex}
\appendix
\section{Categories and Descriptions}
\label{appendix:categories}
See Table \ref{tab:templates}.
\begin{table*}[h]
    \centering
    \begin{tabular}{|p{0.15\linewidth}|p{0.25\linewidth}|p{0.40\linewidth}|p{0.09\linewidth}|}
    \hline
    \textbf{Category} & \textbf{Templates}  &  \textbf{Examples} & \textbf{Instances in Seed Sets}\\
    \hline
    Relative sizes and shapes & Biggest Object, Heaviest Object, Relative Fit &Which of these objects is the lightest? outlet, broom, pail, orange, screen  & 20\\
    &Ease of Interaction Given Object State &Is a raised or lowered drawbridge more effective at getting cars across the river?& \\
     & & Would a shoe fit in a bag? & \\
    \hline
    Object Functions& Basic Affordance, Size Restricted, Shape Restricted, General Property Restricted,  &Which of the following can be used to climb and is bigger than a table? stile, stairway, stepladder, step, ladder &25\\
    & Goal Restricted & What should I use if I want to learn something from the internet? & \\
    \hline
    Object Differences and Hypernyms & Difference within Affordance, Difference within Affordance given Criteria, & What is the difference between a window and a pane? & 25\\ & Basic Is-A, Identical Usage, Sub-Types & Can you use a shed as a barn?&  \\
    & & Choose the truck from the list: coupe,  minivan, 18 wheeler, sedan, ATV& \\
    \hline
    Objects in Risky Situations & Cause Injury, Cause Danger, Cause Object Damage & Which of the following objects would be the most dangerous if it hit something? dvd, screen, wall, drum, mat& 15\\
    \hline
    Required Equipment & How to Use, Equipment for Scenarios, Role of Equipment in Task & Give a step by step explanation of how to use a concrete saw. & 15\\
    & &  What role does a thermal imaging camera play in identifying survivors? & \\
    \hline
    Primary and Secondary Object Facts & Where Object Found, Objects in Location, Secondary Uses & Which of the following can be used as a lever? art, motorcycle, picture, dvd, broom& 15 \\
    \hline
    Disaster Specific Knowledge& Earthquake knowledge &Choose the relevant precautions one should take to prepare for an earthquake.& 5\\
    \hline
    Instruction Following & Instruction Identification, Follow-Up Questions &Choose the navigation instruction: drink from the bottle, sail a boat, enter the doorway & 11\\
    \hline
    \end{tabular}
    \caption{An overview of the types of templates within each category, some examples of resulting seed sentences within each category, and the number of instances of each category within the resulting seed dataset. Note the emphasis on affordances, object knowledge, and instruction knowledge.}
    \label{tab:templates}
\end{table*}
\section{Fine Tuning Specifics}
\label{sec:ft}
For fine-tuning, we used Huggingface TRL\cite{trl} supervised fine-tuning example script modified to access our custom dataset. We used random sampling to split each dataset 90-10 into training and development subsets. We fine-tuned using PEFT \cite{peft} and LORA \cite{lora} to both decrease the computational load on the robot and the time spent fine-tuning. We mostly used parameters suggested by the fine-tuning software we used \cite{trl}, with a learning rate of 2.0e-4, and lora r and alpha values of 32 and 16, respectively. The main differences between our training and the default parameters were training over 3 epochs instead of 1 and not using data packing. We fine-tuned on 2 A100 GPUs.
\section{Synthetic Data Generation Prompting}
We primed Gemini with a system prompt that read as follows:
\begin{displayquote}
    You will be creating multiple choice questions on a variety of topics related to common sense and/or earthquake knowledge. Be creative in choosing the vocabulary and phrasing of these questions. All responses must be given as json objects with the following format: 
    
    \{``instruction'':``example instruction'', ``input'':``A) this B) is C) an D) example E) question'',``output'':``E) Question''\}
\end{displayquote}
A subsequent template prompt from each template category can be seen in Table \ref{tab:prompts}. The corresponding 5 shot examples followed these prompts.
\begin{table*}[h]
    \centering
    \begin{tabular}{|p{0.1\linewidth}|p{0.6\linewidth}|}
    \hline
    Category & Prompt \\
    \hline
    Heaviest & Create 40 unique multiple choice questions about which objects weigh the most. These questions must be multiple choice and they must have 5 options with 1 correct answer. Choose lots of different objects that people interact with.\\
    \hline
    Affordances and Shape & Create 40 unique multiple choice questions about which objects can complete a given function and are a certain shape. \\
     &  These questions must be multiple choice and they must have 5 options with 1 correct answer. Choose lots of different objects that people interact with.\\
     \hline
    Use As & Create 40 unique multiple choice questions about if an object can be used as a substitute for another object. \\
     & These questions must be multiple choice with the two choices being ``it can'' or ``it cannot''. Choose lots of different objects that people interact with. \\
     \hline
     Damage to Objects& Create 40 unique multiple choice questions about which object would cause the most damage to a larger object or structure. \\
      & These questions must be multiple choice and they must have 5 options with 1 correct answer. Choose lots of different objects that people interact with.\\
      \hline
    Equipment Used in Task & Create 40 unique multiple choice questions about how an object is used in a task. The tasks and objects should be related to earthquakes. The answer choices should be brief descriptions of potential ways to use the object in the task. These questions must be multiple choice and they must have 5 options with 1 correct answer. Make sure each answer option is unique. \\
    \hline
    Secondary Uses & Create 40 unique multiple choice questions about objects that are not created to complete a task, but nevertheless can complete the task. These questions must be multiple choice and they must have 5 options with 1 correct answer. \\
    & Make sure the answer choices do not include objects that are meant to do the task described. Make sure to pick lots of unique tasks and objects. \\
    \hline
    Earthquake & Create 40 unique multiple choice questions about earthquakes, earthquake preparation, and earthquake search and rescue protocols. These questions must be multiple choice and they must have 5 options with 1 correct answer. Be as creative as possible with the types of questions you generate, as long as they have something to do with earthquakes.\\
    \hline
    Instruction ID & Create 40 unique multiple choice questions about the purpose of instructions. These questions must be multiple choice and they must have 5 options with 1 correct answer. The answer choices must all be simple instructions. Make sure the correct answer falls under the given category. Use lots of different simple instructions. \\

    \hline
    \end{tabular}
    \caption{A selection of prompts used to generate the synthetic data using Gemini Flash 1.5. Note all prompts had similar language encouraging creativity and strict multiple choice answer requirements.}
    \label{tab:prompts}
\end{table*}
\section{Licenses}
We used TRL \cite{trl} under the Apache License. SemScore \cite{semcode} implements the MIT license, and the LLaMa models were used after author agreement to the LLaMa 3.1 and 3.2 Community License Agreement \cite{llama3}. Ministral 8B Instruct was used under the Mistral Research License \cite{mistral7battention}. 
\end{document}